\documentclass{article}

\usepackage{microtype}
\usepackage{graphicx}
\usepackage{subfigure}
\usepackage{booktabs} 
\usepackage{amssymb}
\usepackage{amsmath}
\usepackage{tabularx}
\usepackage{makecell}
\usepackage{float}
\usepackage{xcolor}
\usepackage{comment}
\usepackage{xspace}
\usepackage{multirow}

\usepackage{hyperref}

\newcommand{\titl}{When does loss-based prioritization fail?}

\graphicspath{{figures/}{drawings/}}

\usepackage{ownstyles}

\usepackage[accepted]{icml2021}

\icmltitlerunning{When does loss-based prioritization fail?}

\begin{document}

\twocolumn[
\icmltitle{When does loss-based prioritization fail?}



\icmlsetsymbol{equal}{*}

\begin{icmlauthorlist}
\icmlauthor{Niel Teng Hu}{mlc}
\icmlauthor{Xinyu Hu}{ub}
\icmlauthor{Rosanne Liu}{mlc,goo}
\icmlauthor{Sara Hooker}{goo}
\icmlauthor{Jason Yosinski}{mlc}
\end{icmlauthorlist}

\icmlaffiliation{mlc}{ML Collective}
\icmlaffiliation{goo}{Google Research, Brain}
\icmlaffiliation{ub}{Uber}

\icmlcorrespondingauthor{Niel Teng Hu}{hu.niel92@gmail.com}

\icmlkeywords{Machine Learning, ICML}

\vskip 0.3in
]


\printAffiliationsAndNotice{}  
\begin{abstract}
Not all examples are created equal, but standard deep neural network training protocols treat each training point uniformly. Each example is propagated forward and backwards through the network the same amount of times, independent of how much the example contributes to the learning protocol. Recent work has proposed ways to accelerate training by deviating from this uniform treatment. Popular methods entail up-weighting examples that contribute more to the loss with the intuition that examples with low loss have already been learned by the model, so their marginal value to the training procedure should be lower. This view assumes that updating the model with high loss examples will be beneficial to the model. However, this may not hold for noisy, real world data. In this paper, we theorize and then empirically demonstrate that loss-based acceleration methods degrade in scenarios with noisy and corrupted data. Our work suggests measures of example difficulty need to correctly separate out noise from other types of challenging examples.



\end{abstract}

\section{Introduction}

Recent years have observed a rapid explosion in the size and training costs of deep neural networks. Training ever larger neural networks exacerbates the cost and time required for training \citep{2018Amodei} and the carbon footprint of the model \citep{strubell2019energy}. Long training times and high cost hinder the democratization of deep learning research and applications from people with limited budgets or few resources \citep{hooker2020hardware,obandoceron2020revisiting}.

However, much of this cost could be avoided with more efficient training. A typical training regime makes the costly choice of treating all examples equally even though the value of the information in each example may not be the same. Assuming all examples are equally important and propagating each forward and backwards through the network the same amount of times results in redundancy and inefficient use of training budget.

Recent work proposes accelerating training by treating data examples differently. \citet{jiang2019accelerating} and \citet{Angelos2018} have proposed loss-based sampling methods to speed up deep neural network training by emphasizing the high loss examples. Although such methods achieve significant acceleration, the merits of each have been evaluated on small curated research datasets that do not necessarily represent the messiness of real world data. In large scale training settings, models are often trained on datasets with unknown quality and degree of input or label corruption \citep{Tsipras2020FromIT,hooker2020compressed,beyer2020imagenet}.

Intuitively, these real-world settings present a particular challenge for loss-based sampling approaches because two distinct categories of examples have high-loss: (a) difficult or low-frequency examples and (b) corrupted, noisy, or mislabeled examples. The former are more useful for training than the median example and are adeptly selected by loss-based approaches. But the latter are less useful than the median. In fact, mislabeled examples hurt training and decrease generalization, and so loss-based methods that systematically boost their salience during training may do more harm than good.

In this paper we evaluate the robustness of loss-based sampling methods to varying levels of dataset noise and corruption. Our goal is to spur a more nuanced discourse about what we mean by hardness of examples and the assumptions motivating loss-based prioritization methods. We define three types of modifications to create artificial corrupted examples: \textbf{(1)} label randomization, \textbf{(2)} pixel shuffling, and \textbf{(3)} replacing inputs with Gaussian noise. We have observed that all these transformations corrupt examples such that a human can no longer identify the target class. Thus, our definition of noise centers on corruptions which introduce sufficient stochasticity that the mapping can no longer be learnt by a human.

We study the performance of two representative loss-based sampling methods, Selective Backprop (SB) \citep{jiang2019accelerating} and the variance reduction importance sampling method proposed in \citep{Angelos2018}, on artificial corrupted datasets and try to explain the intuition behind the failing cases. We find that acceleration of these methods derails under supervised learning tasks with corrupted or noisy information, however the variance reduction importance sampling method didn't degrade as severely as SB. We show that for both methods considered, the degradation to performance occurs due to the upweighting of these high-loss, out-of-distribution examples.


\textbf{Implications of work.} Loss-based acceleration methods present the attractive proposition of being able to accelerate any loss-based training method, but as we show in this paper, this assumes that the highest loss examples are informative for the task at hand. Often, the source of noise in challenging examples is irreducible. Even with an up-weighting strategy, a useful mapping between input and output space cannot be learnt. Real world datasets contain varying degrees and types of corruption, and as we show here, corruption causes these acceleration methods to degrade. This suggests that either better acceleration methods are needed or that rigorous data pre-processing to remove corruptions should be applied before deploying these methods. A rich subject of future research is developing estimation methods that distinguish between noisy examples which can derail training and the more informative atypical examples. 





\section{Loss-based Sampling Methods}

In a supervised learning setting, let $x_i, y_i$ be the i-th example of the training set where $x_i$ represents the input tensor to the network and $y_i$ represents the label. Let $f_{w}$ be the neural network model parameterized by learnable parameter $w$ and $L$ the cross-entropy loss. The goal of training is to find
    
\begin{center}
$w^{*} = \underset{w}{\mathrm{argmin}} \frac{1}{N} \sum_{i=1}^{N} L(f_w(x_i), y_i)$
\end{center}

Where $N$ represents the number of examples in the training set. In mini batch SGD, we uniformly sample a batch of examples from the dataset without replacement and use the average gradient from these examples to update the parameters, here using a learning rate $\eta_t$.

\begin{center}
$w_{t+1} = w_{t} - \eta_t  \frac{1}{N} \sum_{i=1}^{N} \nabla_{w} L(f_w(x_i), y_i)$
\end{center}

As may be seen in the equation above, the model weights all training examples equally. In contrast, the loss-based sampling methods that we evaluate up-weight challenging examples. We introduce both methods below:
    
\textbf{Selective Backprop \textbf{(SB})}. SB  \citep{jiang2019accelerating} is a framework proposed to prioritize learning high loss examples every iteration. In the original paper, SB converges to target error rates up to 3.5x faster than standard SGD and can be further accelerated by using stale forward pass information.

SB works by maintaining a moving histogram of size $H$ and a buffer for candidate examples. At each iteration, SB computes forward passes to calculate the loss of each example. Then, for each example, its loss is input to a function which outputs the probability that it should be sampled. Samples are then chosen and pushed into the candidate buffer. When the buffer size exceeds the batch size $B$, the first $B$ examples are used to compute the model gradient updates.

The probability of each example being sampled is calculated by the CDF of its loss from the histogram.
\begin{center}
$P(L(f_w(x_i), y_i)) = CDF(L(f_w(x_i), y_i))^{\beta}$
\end{center}

$\beta >= 0 $ is a hyper-parameter controlling the selectivity of the algorithm. The percentage of examples selected in a batch $\approx \int_{0}^{1} p^{\beta}  = 1 / (\beta + 1)$, so $\beta$ = 0 corresponds to 100\% selection (regular SGD), $\beta$ = 1 corresponds to 50\% selection with linearly ramping probability from 0\% probability of the minimum loss example being chosen to 100\% probability of the highest loss example being chosen, and $\beta$ = 2 corresponds to 33\% selection with a similarly arranged quadratic ramp.
 
\textbf{Variance Reduction Importance Sampling (\textbf{VR})}. We also consider the importance sampling approach of \cite{Angelos2018} using loss values. The paper proposed an importance sampling scheme that prioritizes computation on examples that reduce the variance of the gradient estimates.

Similarly to SB, the method maintains a pool of pre-sampling candidate examples. During training, mini-batches are pushed into this pool and once the size of pool exceeds a predefined size $B$, the algorithm samples examples from a distribution proportional to their loss values. The size of the pre-sampling pool $B$ is an additional hyper-parameter introduced. The paper also derives an estimator of the variance reduction and switches importance sampling on when it is estimated to produce a speedup.

In this paper, we test specifically the loss-based, no up-weighting, no warm up version of importance sampling from \cite{Angelos2018}, because this is the version adopted by \cite{jiang2019accelerating} in their tests.

\begin{table*}[t]

\vskip 0.15in
\begin{center}
\small
\begin{sc}
\begin{tabular}{c|ccccccc}
\toprule
\makecell[r]{\textbf{Corruption $\rightarrow$}} & 
  \multicolumn{1}{c}{Pristine} & 
  \multicolumn{2}{c}{Random Labels} & 
  \multicolumn{2}{c}{Shuffle Pixels} &
  \multicolumn{2}{c}{Gaussian Generated} 
  \\ \hline
  \textbf{$\downarrow$ Algorithm} & \textbf{0\%} &
  \textbf{25\%} & \textbf{50\%} & 
  \textbf{25\%} & \textbf{50\%} &
  \textbf{25\%} & \textbf{50\%}
\\ \midrule

  Standard & 1x  & 1x   & 1x  & 1x  & 1x 
  & 1x  & 1x 
  \\
   & (4.28 $\%$) & (8.53 $\%$) & (13.27 $\%$) & (4.98 $\%$) & (6.88 $\%$) & (5.27 $\%$) & (6.72 $\%$)
  \\
  
  SB (50$\%$ selectivity) & 2.0x  &  2.0x  & 1.5x  & 2.0x  & 2.0x & 2.0x  & - 
  \\
  & (4.26 $\%$) & (9.03 $\%$) &  (14.66 $\%$) &  (5.48 $\%$) & (7.71 $\%$) & (5.68 $\%$) &  (8.47 $\%$)
  \\
  
  SB (33$\%$ selectivity) & 3.0x  & -  &  - 
  & -  & -  & -  & - 
  \\
  & (4.39 $\%$) & (12.28 $\%$) &  (33.73 $\%$)
  & (6.31 $\%$) & (9.95 $\%$) & (7.18 $\%$) & (17.92 $\%$)
  \\
  
  VR (max 33\% selectivity) & 1.7x & 1.0x  & 1.0x & 1.1x & 1.0x & 1.0x & 1.0x
  \\
  & (4.81 $\%$) & (8.65 $\%$) & (13.20 $\%$) &  (5.71 $\%$) & (6.95 $\%$) & (5.35 $\%$) & (6.89 $\%$)
  \\
  
  VR (max 50\% selectivity) & 1.8x & 1.0x & 1.0x & - & 1.1x & 1.1x & 1.0x
  \\
  & (4.87 $\%$) & (8.74 $\%$) & (13.14 $\%$) & (6.27 $\%$) & (7.69 $\%$) & (6.30 $\%$) & (6.70 $\%$)
  \\

\bottomrule
\end{tabular}
\end{sc}

\end{center}
\vskip -0.1in

\caption{Speedup and error rate of various methods under different datasets.}
\tablabel{table:speedup}

\end{table*}

\section{Experiments}

We evaluate the SB and VR methods with a standard image classification dataset, CIFAR 10 \citep{Krizhevsky09}, as well as corrupted variations of it. We first explain how we create noisy examples.

\subsection{Creating Noisy examples}

\newcommand{\dtrain}{\ensuremath{\mathcal{D}_{\mathrm{train}}}\xspace}
\newcommand{\dtest}{\ensuremath{\mathcal{D}_{\mathrm{test}}}\xspace}

We denote the clean training and test datasets as 
\dtrain and \dtest respectively and $f_{w}$ be the neural network model trained on \dtrain where $w$ are the learnable parameters. The objective is to apply modifications on \dtrain where $\dtrain = \{(x_i, y_i)\}^n_{i=1}$ with inputs $x \in \mathcal{X}$ and output $y \in \mathcal{Y} = \{1, ..., K \}$ where $K$ is the number of classes.

We consider the following modifications on example $(x_i, y_i)$ to artificially introduce noise:

1. {\bf Random labels}: Output $y_i$ is replaced by $y_i'$ which is uniformly sampled from the set $\{1, ..., K \}$ including the original $y_i$ label.

2. {\bf Shuffled pixels}: A single random permutation $p$ is chosen for the task and is applied to turn each $x_i$ into a permuted version $x_i'$.

3. {\bf Gaussian}: $x_i$ is replaced by a $x_i'$ in which each pixel is sampled from a Gaussian distribution whose mean and variance match the mean and variance of pixel values in $x_i$.

These noise-inducing methods, while artificial, attempt to recreate the real-world scenario where, for example, label error is commonly seen~\citep{snowetal2008cheap,yan2014} or hardware collection noise adds artefacts to a subset of the data.


\subsection{Evaluating loss-based sampling}

We evaluate using CIFAR10, which contains 50000 training images and 10000 test images, divided into 10 classes respectively, and each example is a 32 $\times$ 32 image with three color channels. We randomly sample $25\%$ or $50\%$ of the dataset and apply the modifications described in the previous subsection.

For each corruption type and percentage, we run 5 variants of acceleration algorithms, SB-0 (Selective Backprop with $100\%$ selection, which is standard SGD), SB-1 (Selective Backprop with $50\%$ selection), SB-2 ($33\%$ selection), VR with max selectivity $50\%$, and VR with max selectivity $33\%$.

We train a Wide Residual Network \citep{zagoruyko2017wide} with \textit{depth} = 28 and \textit{widen\_factor} = 10. We use \textit{batch\_size} = 128, \textit{lr} = 0.1 and \textit{momentum} = 0.9. We use a standard SGD optimizer with \textit{weight\_decay} = 0.0005.  The learning rate drops by $80\%$ at epoch 60 and 80 and the training runs for 100 epochs. For each training image, first we crop the given image at a random location with padding on the borders, then horizontally flip the image with $50\%$ probability, and lastly normalize the image. For the test set, we only normalize the dataset without data augmentation.  We ran the experiment with 5 random seeds and average the results.
 
In this paper, consistent with prior work \citep{jiang2019accelerating}, we use \textbf{the number of examples back-propagated to reach a target test error} as a proxy measure for speedup. We record the best test error of standard non-accelerated training as the target and when measuring the speedup of a acceleration method, we use the number of back-propagation to reach the 1.2x non-accelerated best test error as a measure of speed. 
  

\subsection{Results}

The results are shown in \tabref{table:speedup}. A test error threshold is chosen by running standard SGD training and multiplying the best error achieved by 1.2. Speedup is the number of back-propagations standard SGD requires to reach this threshold divided by the number required by the given method. Dashes indicate the network was unable to reach the threshold error. Numbers in parentheses indicate the best test error achieved by the method averaged over five runs. With 0\% corruption both SB and VR significantly accelerate training. When corruption is applied to the data, both methods either fail to deliver a speedup or attain a worse test error.
  
We find that on the pristine (un-corrupted) CIFAR10 dataset, both SB and VR are able to reach the target test error with speedup. As we increase the amount of corruptions in the dataset, the speedup degrade compared to pristine datasets and both algorithms converge to higher test error. Training plots are included in the supplementary material.

Noticeably, VR didn't degrade as severely as SB when more corruptions are introduced into the datasets. \citet{Angelos2018} show that the variance reduction is proportional to the squared $\textit{L2}$ distance between the sampling distribution and uniform distribution. By tracking the squared distance $\textit{L2}$, VR only enables importance sampling when there is a guaranteed variance reduction. We believe in our experiments with artificial corrupted examples, VR cancelled the importance sampling most steps which leads to only minor degradation vs. standard training but does not produce a speedup.

\begin{figure}[H]
    \centering
    \includegraphics[width=1\linewidth]{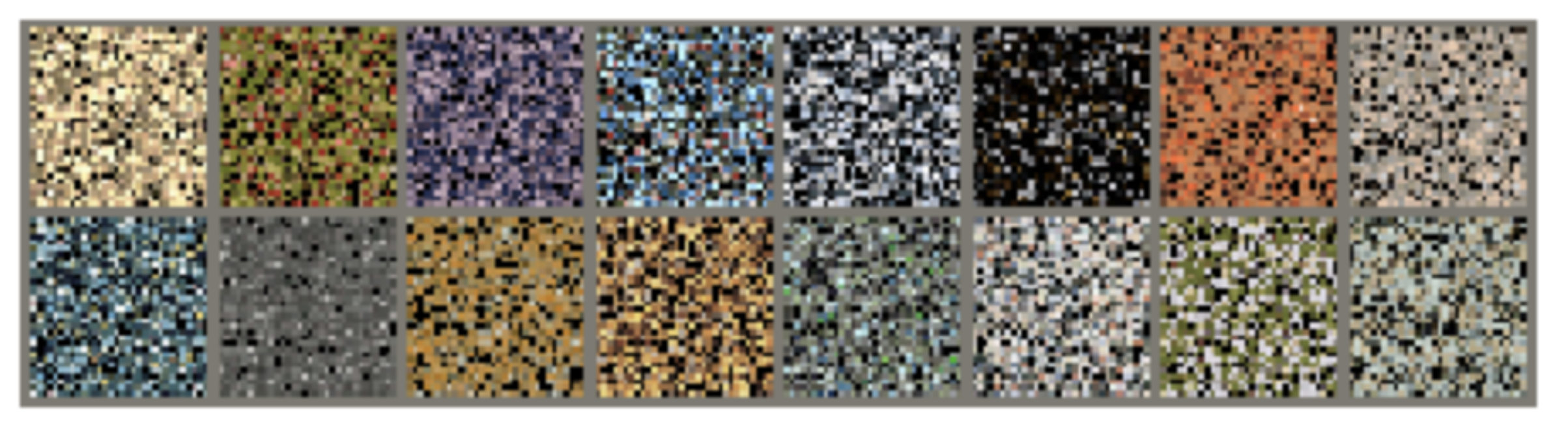} 
    \\
    \includegraphics[width=1\linewidth]{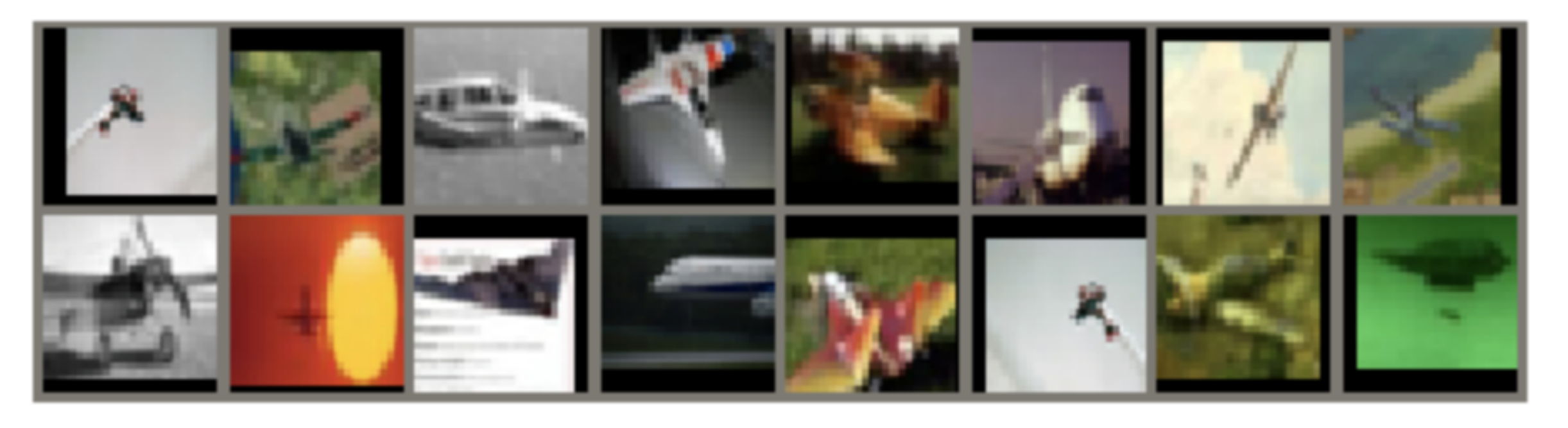}
    \caption{16 images sampled from the 2000 most frequent picks by Selective-Backprop on \textbf{(top)} $25\%$-shuffle-pixel CIFAR10 vs. \textbf{(bottom)} the pristine dataset. SB tends to prioritize noisy images instead of the difficult (high-loss) noiseless examples that would lead to accelerated, generalizable learning.
    }
    \label{fig:comparison}
    \vspace{-.3em}
 \end{figure}

 \section{Discussion}
 
{\bf Why the degradation in the corrupted datasets.} We hypothesize that the degradation is caused by these modified examples being prioritized by the acceleration method. The randomization of label deliberately destroys the links between inputs and target labels therefore contains no value for generalization. For Gaussian generated and pixels shuffled images, the modification transforms the inputs into noise-like images while only preserving some global statistics. These modifications eliminate structures in the images and create out-of-distribution examples that don't contribute to generalization. 
 
\begin{figure}[h]
\includegraphics[width=1\linewidth]{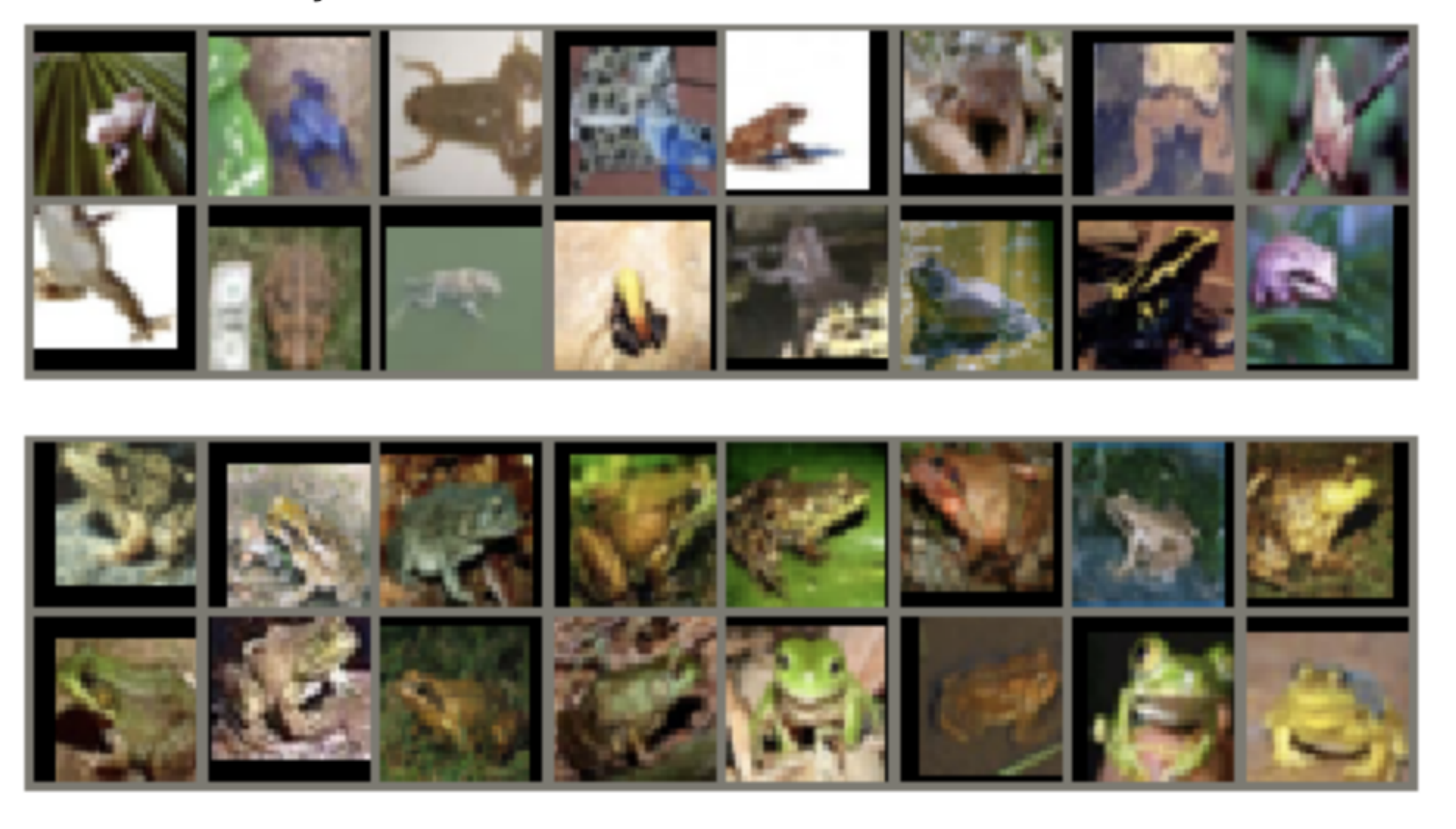}
\includegraphics[width=1\linewidth]{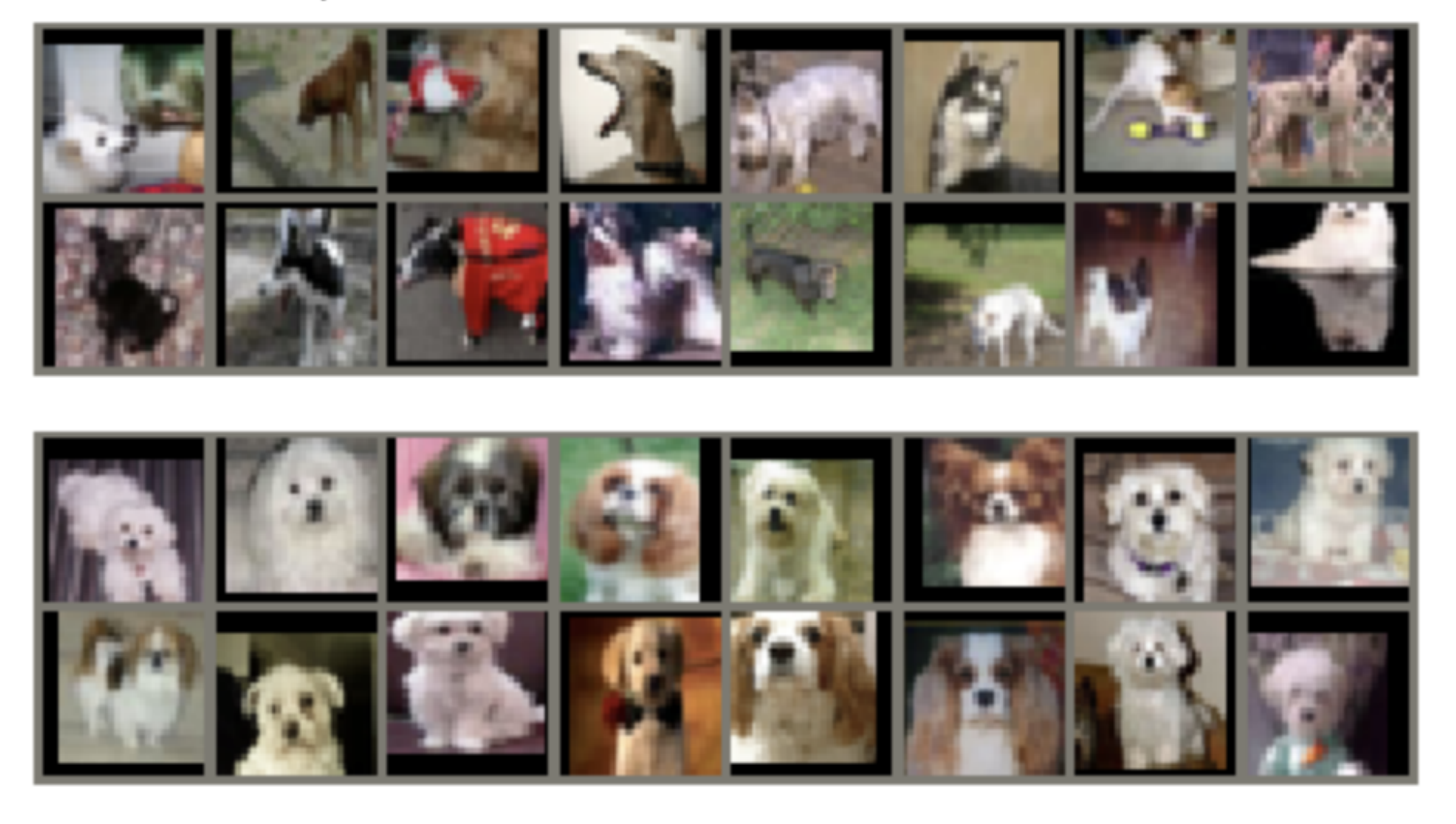}
\caption{Images sampled by SB and VR on the pristine CIFAR10 dataset.
  Shown are 16 images sampled from the 2000
  \textbf{(first row)} most frequent picks by SB,
  \textbf{(second row)} least frequent picks by SB,
  \textbf{(third row)} most frequent picks by VR, and
  \textbf{(fourth row)} least frequent picks by VR for a given single class.
  The left column shows images from the ``frog'' class and the right shows images from the ``dog'' class.
}
\label{fig:top_images}
\vspace{-.3em}
\end{figure}

We visually examine the top picked examples from both datasets. In \figref{comparison}, we sample 16 images (belong to one class) from the 2000 most frequent picks by SB on modified and unmodified datasets. In \figref{top_images}, we sample 16 of the most frequent and least frequent picks by SB for two classes (frogs and dogs).

It is observed that examples less likely to be picked often share similar visual characteristics: similar shape, contours, color or image composition. Their similarity makes them more redundant during training, so some may be dropped without damaging the training process.
On pristine datasets, those examples more likely to be picked show more diverse compositions. Unfortunately, when datasets are corrupted, those corrupted examples tend to be chosen.

\begin{figure*}[t]
\centering
\includegraphics[width=0.33\linewidth]{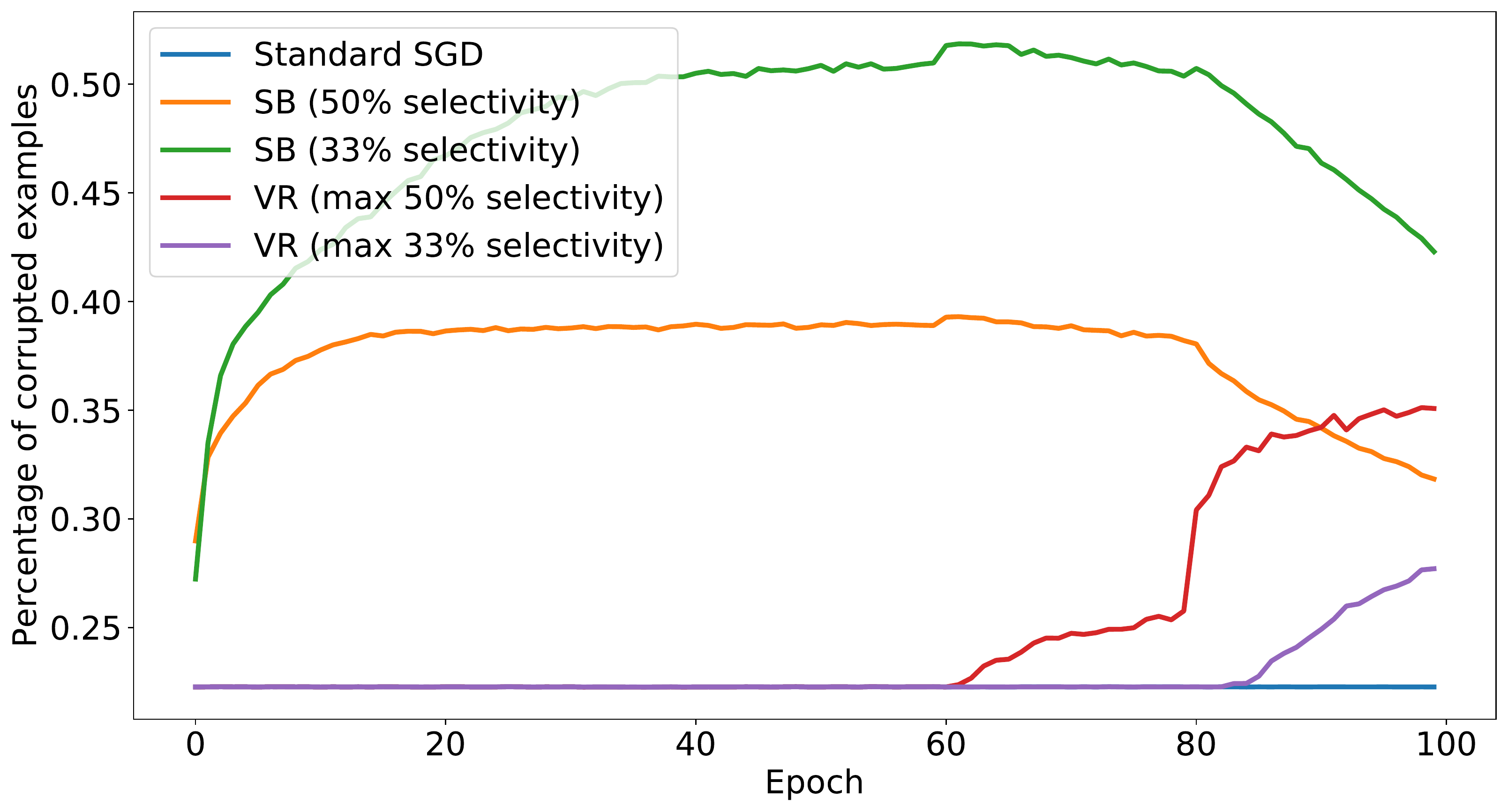}
\includegraphics[width=0.33\linewidth]{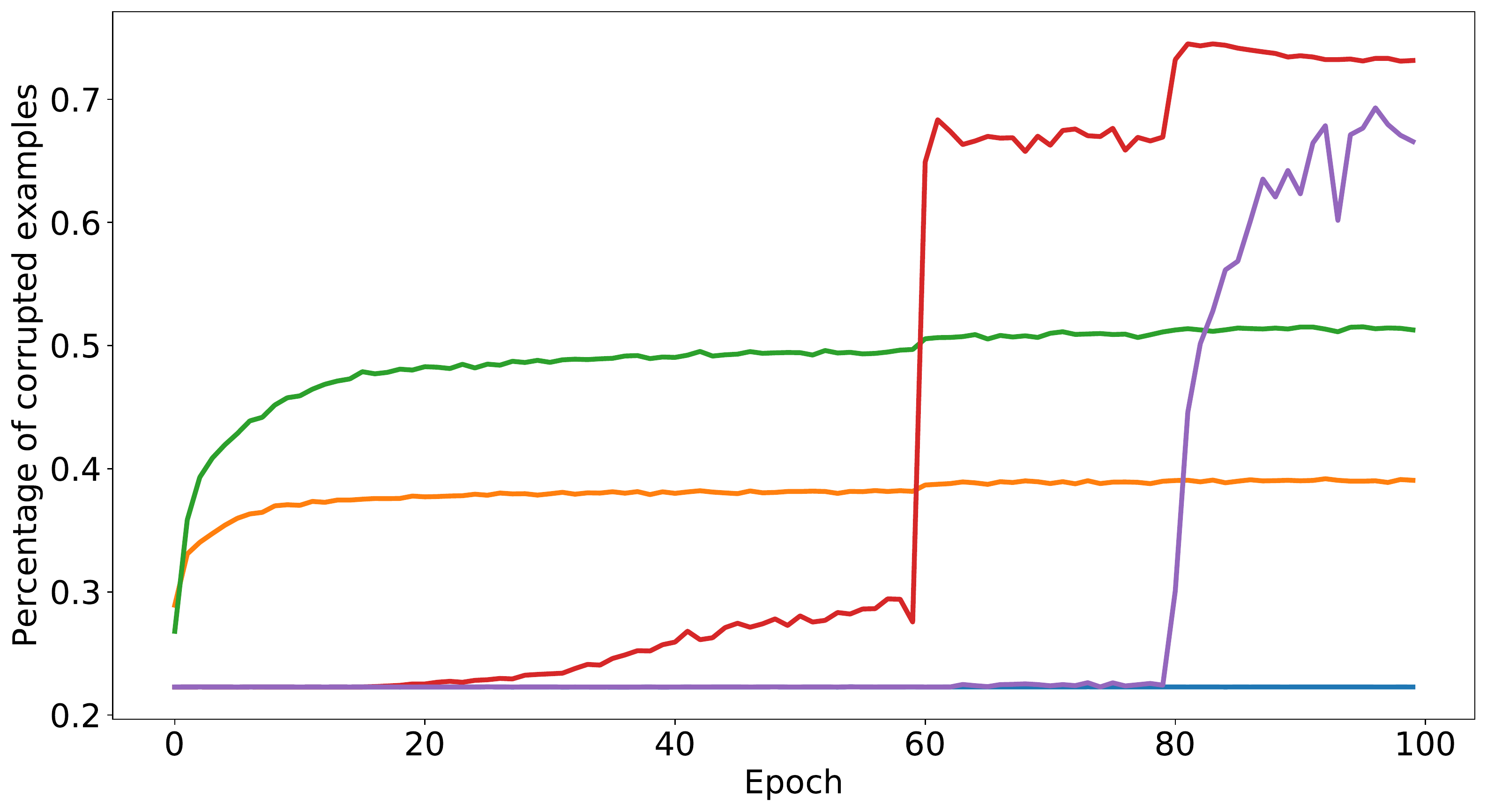}
\includegraphics[width=0.33\linewidth]{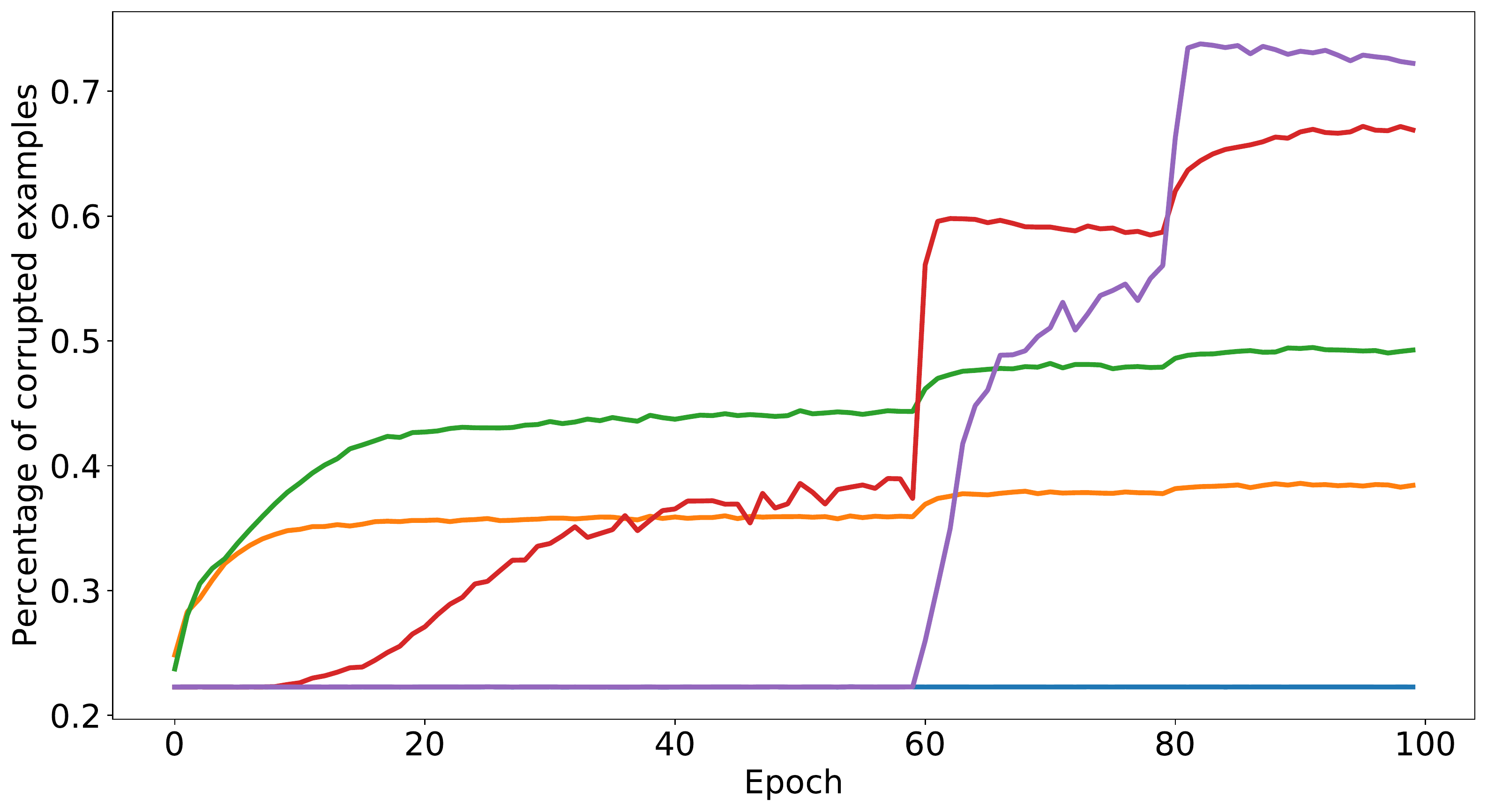}
\caption{Percentage of corrupted examples selected for inclusion in CIFAR10 training batches where 25\% of images are corrupted via
  \textbf{(left)} random labels,
  \textbf{(middle)} Gaussian generated pixels
  \textbf{(right)} shuffled pixels.
  Both SB and VR sample corrupted examples more often than clean examples, which degrades training performance.
}
\figlabel{noise_pct}
\vspace{-.3em}
\end{figure*}


We hypothesized that for corrupted datasets, both atypical examples and corrupted examples produce high losses, leading the acceleration method to prioritize both. To validate this hypothesis, we plot the percentage of corrupted examples included in each training iteration (\figref{noise_pct}) and find that across different type of corruptions, both SB and VR consistently over-sample corrupted examples compared to the constant percentage of standard training. The results suggest that, in contrary to the promises of loss-based acceleration methods, these methods might actually hurt the training in partial corrupted datasets.

\begin{figure*}[ht!]
    \centering
    \includegraphics[width=0.33\linewidth]{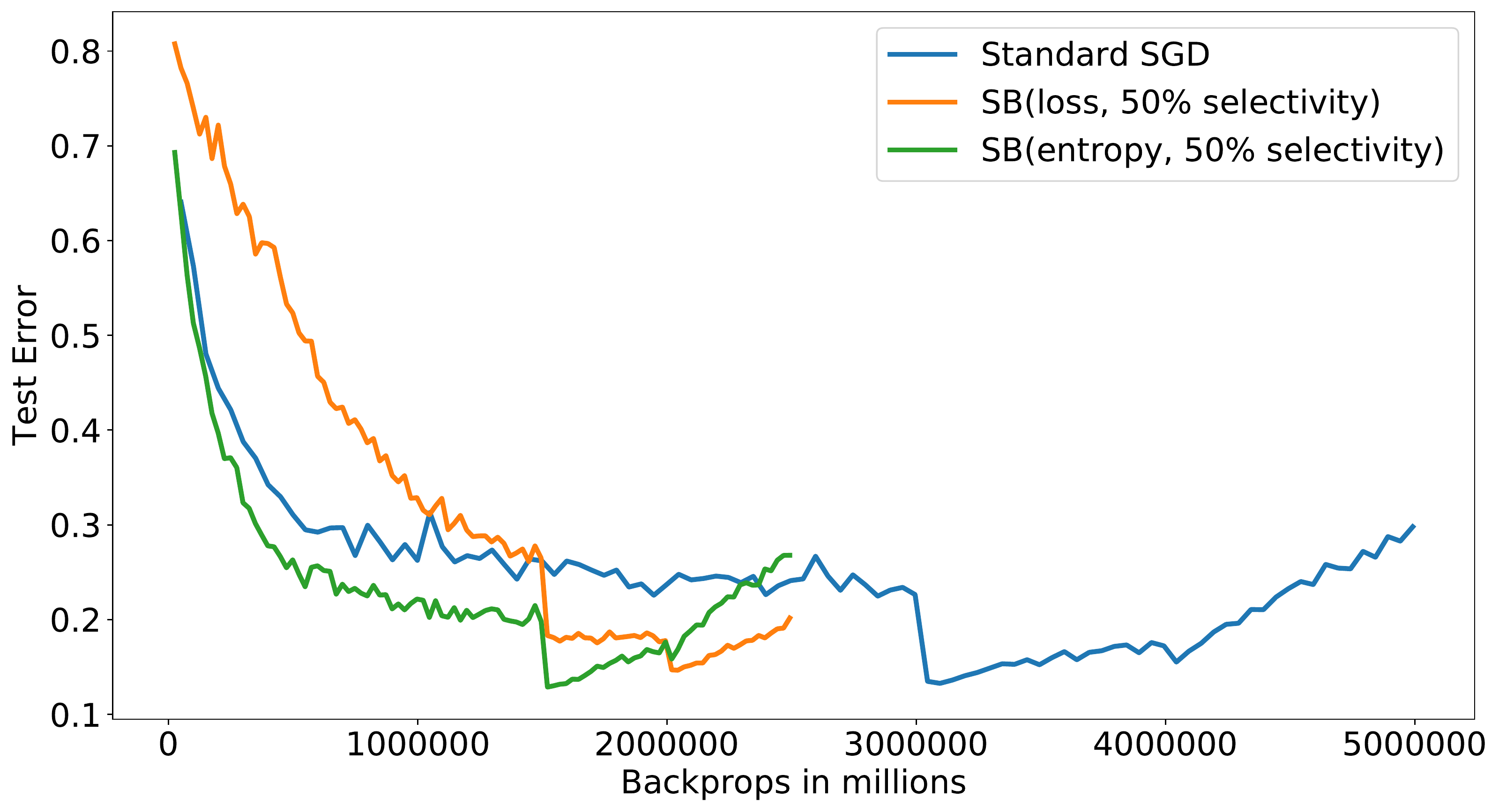}
    \includegraphics[width=0.33\linewidth]{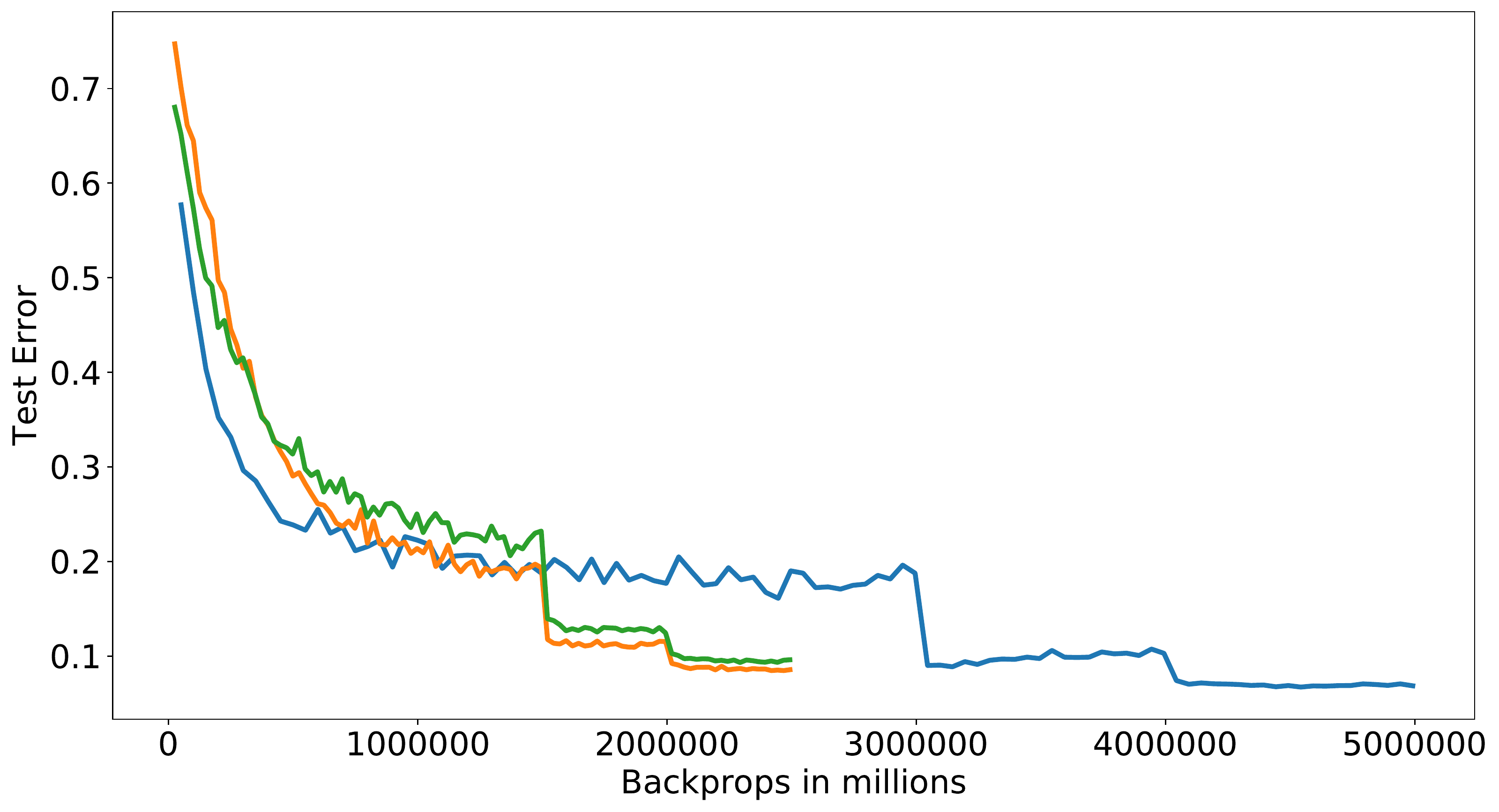}
    \includegraphics[width=0.33\linewidth]{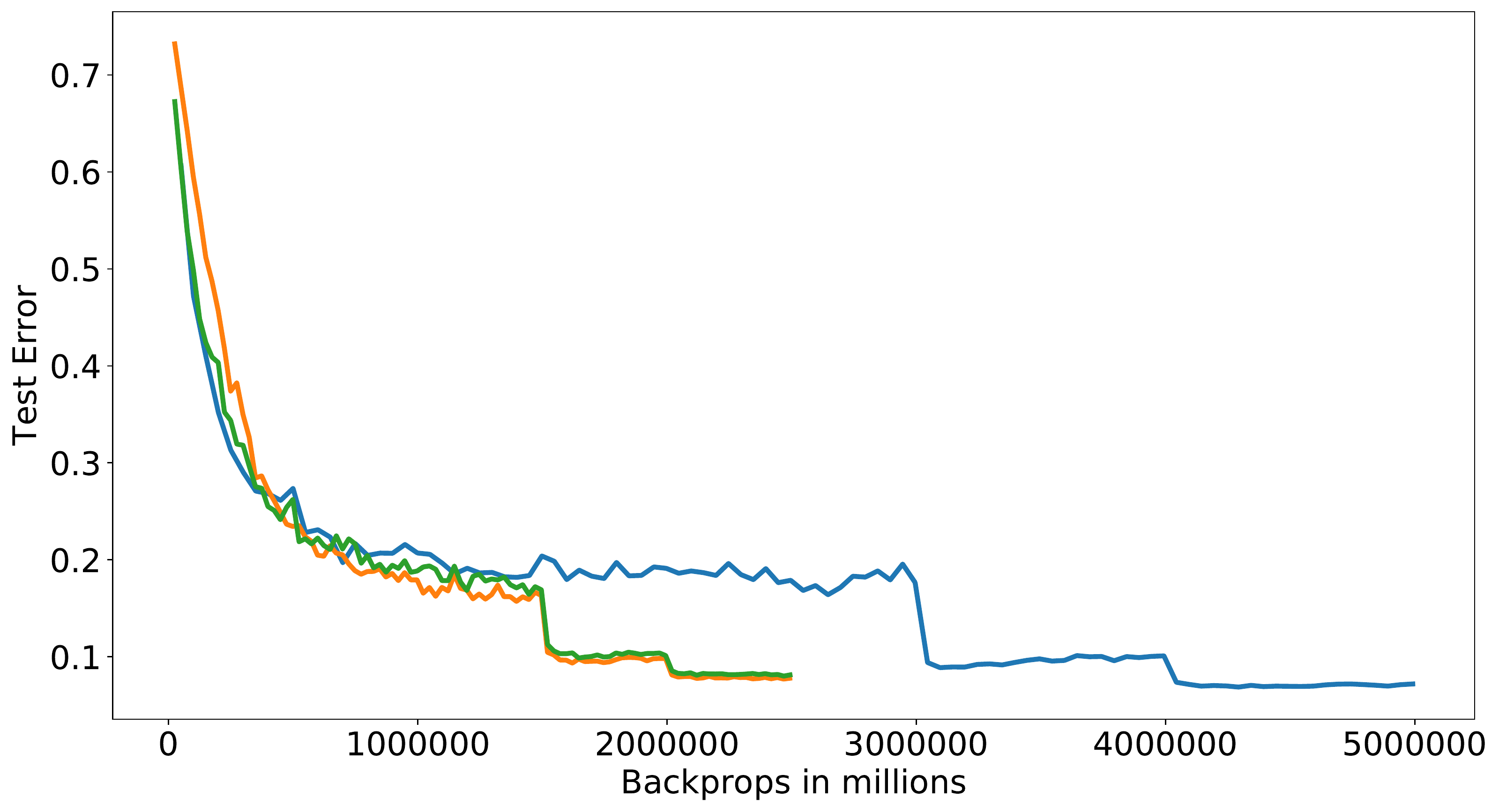}
    \caption{Test error over the course of training for SB with entropy prioritization target vs. vanilla SGD and SB with Loss. Results are shown on CIFAR10 with
      \textbf{(left)} 50\% random labels,
      \textbf{(middle)} 50\% Gaussian generated images, and
      \textbf{(right)} 50\% images with shuffled pixels.
    }
    \figlabel{SBE}
    \vspace{-.3em}
 \end{figure*}

{\bf Entropy-Based Selective Backprop (SBE).}
In an attempt to fix SB's vulnerability, we switch the prioritization target from the cross-entropy loss to the model's uncertainty, calculated as the entropy of the prediction distribution. Intuitively, entropy-based sampling encourages the model to learn those examples that cause the most uncertainty over multiple classes.

The method works similar to SB except at each iteration, SBE computes forward passes and calculates the entropy of the prediction distribution for every example and then updates the moving histogram. The probability of each example being sampled is calculated by the CDF of its entropy with respect to the histogram. Samples are then chosen and pushed into the candidate buffer. When the buffer size exceeds the batch size $B$, the first $B$ examples are fetched to compute the gradient.

As shown in \figref{SBE}, we find that this approach improves training significantly in the face of random labeling corruptions, but it performs worse than loss-based SB when datasets contain Gaussian generated examples.
In CIFAR10 with 50\% random labels, entropy-based SB is able to achieve the 1.2x standard test error with 2.0x speedup. In CIFAR10 with 50\% Gaussian generated examples, entropy-based SB performs slightly worse than loss-based SB.

\section{Related Work}
Many methods have been proposed to improve neural network training. Curriculum learning \citep{Bengio2009-pr} devised a strategy of supplying easy and prototypical examples first and gradually increases their difficulty, which has shown to be beneficial to the overall generalization of the model. In real-world applications where identification of easy/hard examples could be difficult, Self-paced learning \citep{Kumar2010-zc} infers the difficulty of examples from their corresponding training loss during training.

Another common approach is to use importance sampling. The basic approach is to over-sample a subset of examples, then to weight them with the inverse of the sampling probability so the gradient estimator is unbiased \citep{Angelos2018, NEURIPS2018_967990de, gao2015active}. Among these works, many \citep{Angelos2018, DBLP:journals/corr/LoshchilovH15, Schaul2016} use the loss to generate the sampling distribution and sample examples proportional to the historical loss. Although most methods require maintaining a data structure proportional to the training set in size, e.g. full history of training loss for each example, or training an extra auxiliary DNN, \citep{Angelos2018} doesn't have the requirement to maintain historical data for each example and therefore may scale to large datasets and to the incremental learning setting. 

Besides, \citep{jiang2019accelerating} proposes a novel framework to prioritize high loss examples and speedup training without maintaining a data structure proportional to the training set in size. \cite{chang2018active} proposes up-weighting examples based on estimates on model uncertainty. \cite{yoon2019data} proposes a meta learning framework which models the value of each example using a deep neural network. 

Mitigating the impact of noisy labels has also been the subject of considerable research \citep{tanaka2018joint}.

\section{Conclusions}

In this paper, we showed that the acceleration from prioritizing high loss examples does not always hold when we cannot guarantee the dataset is high quality. We showed the degradation of using loss-based acceleration through experiments with artificially corrupted datasets. In our experiments, two acceleration algorithms actually hurt the training by over-sampling the corrupted examples, which confirmed our hypothesis that loss is an imperfect estimate of how challenging an example is. Up-weighting based upon loss is not robust in the face of irreducible noise.



\bibliography{main}
\bibliographystyle{icml2021}


\appendix
\clearpage

\renewcommand{\thesection}{S\arabic{section}}
\renewcommand{\thesubsection}{\thesection.\arabic{subsection}}

\newcommand{\beginsupplementary}{%
    \renewcommand{\thetable}{S\arabic{table}}%
    \renewcommand{\thefigure}{S\arabic{figure}}%
}
\beginsupplementary

\onecolumn

\begin{center}
    {\LARGE\sc Supplementary Information for:\\ \titl\par}
\end{center}

\section{Additional plots}

\begin{figure*}[ht!]
\vskip 0.2in
\centering
    \begin{subfigure}
    		\centering
        	\includegraphics[scale=0.35]{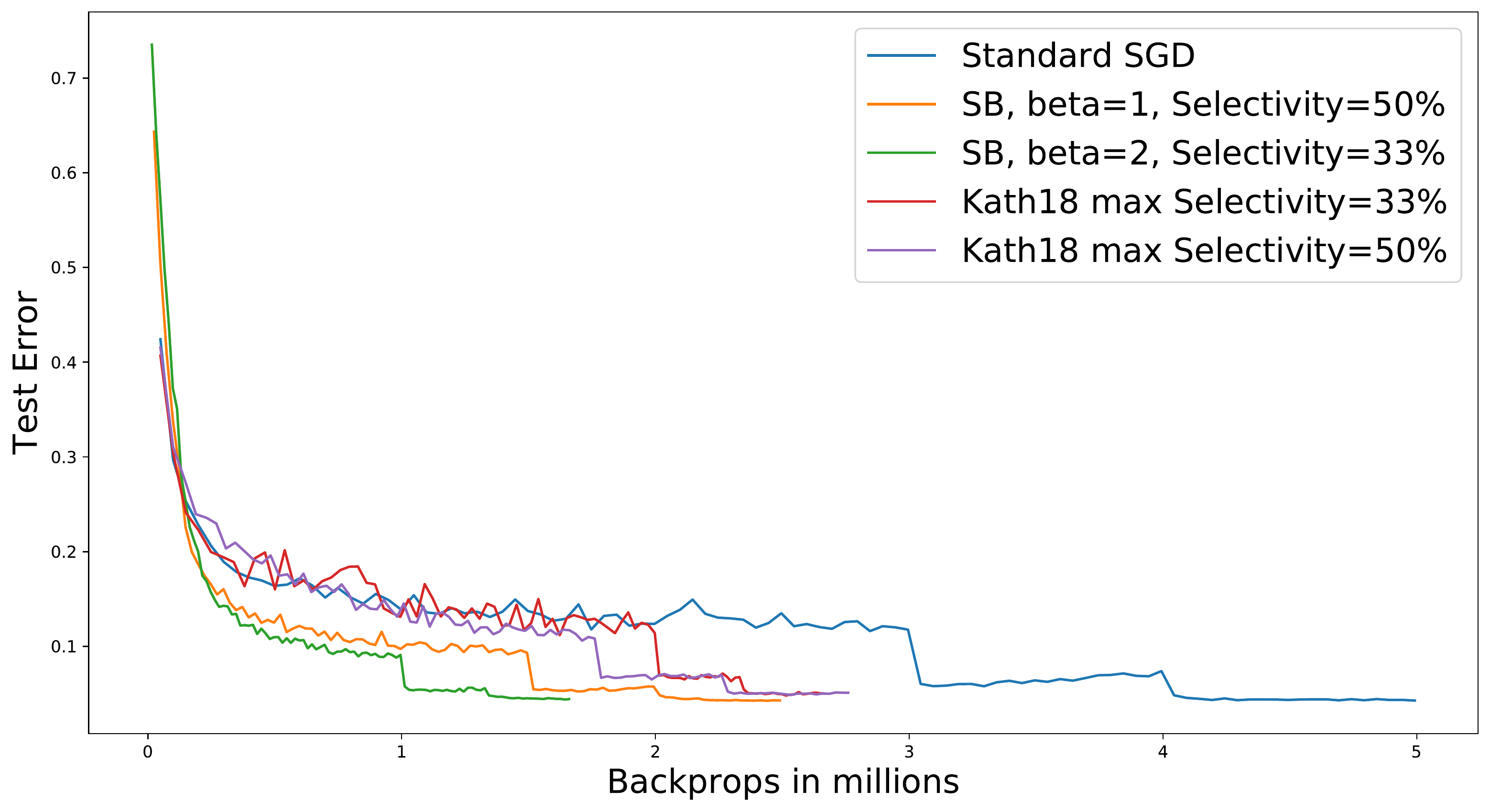}
        	\caption{No Dataset Corruption} \label{fig:no corruption}
    \end{subfigure} 
    \begin{subfigure}
		\centering
    	\includegraphics[scale=0.35]{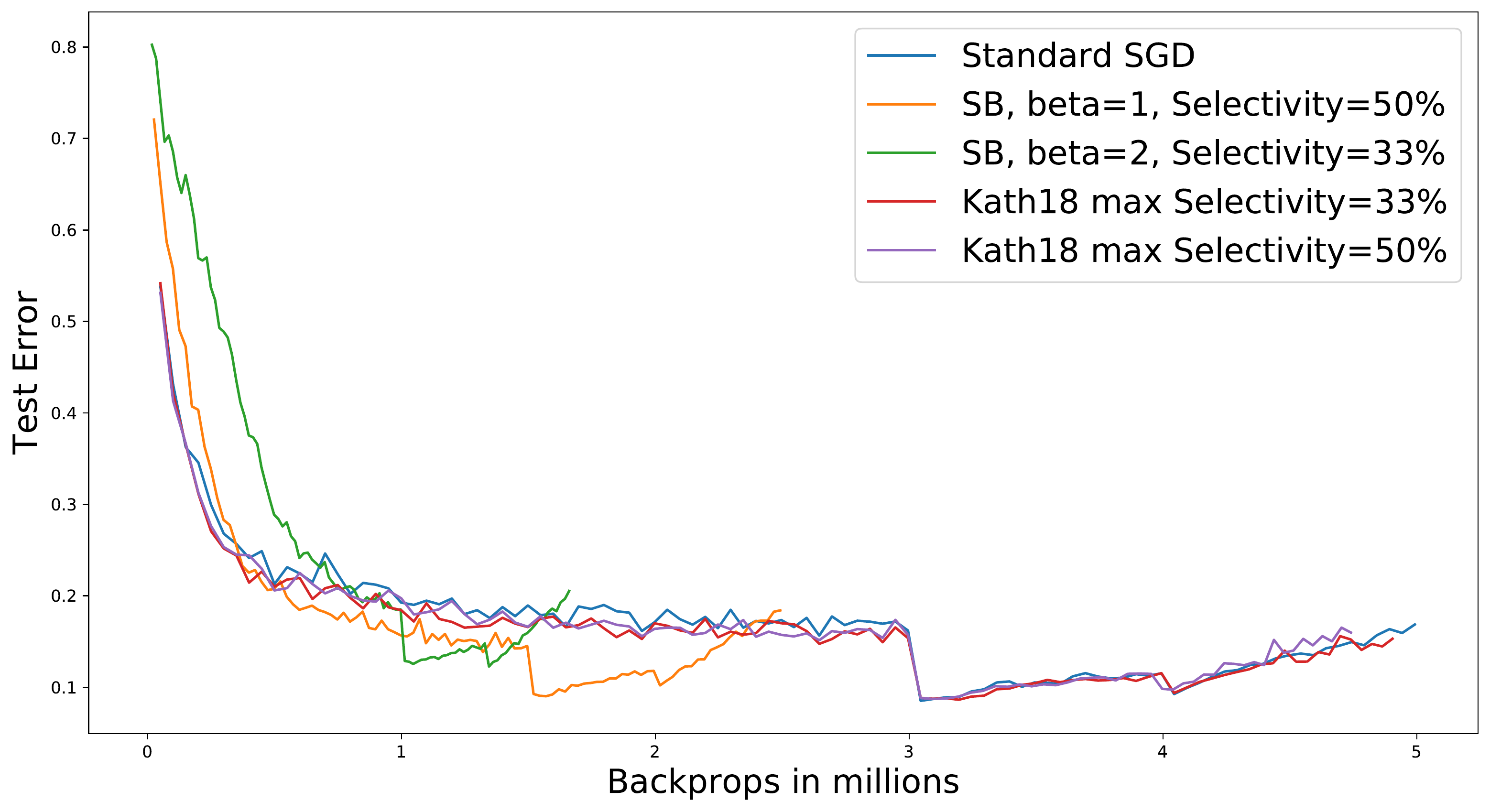}
    	\caption{Corrupted Labels (25 $\%$) \label{fig:25noise}}
	\end{subfigure} 
\end{figure*}

\begin{figure*}[ht!]
\centering
	\begin{subfigure}
		\centering
    	\includegraphics[scale=0.35]{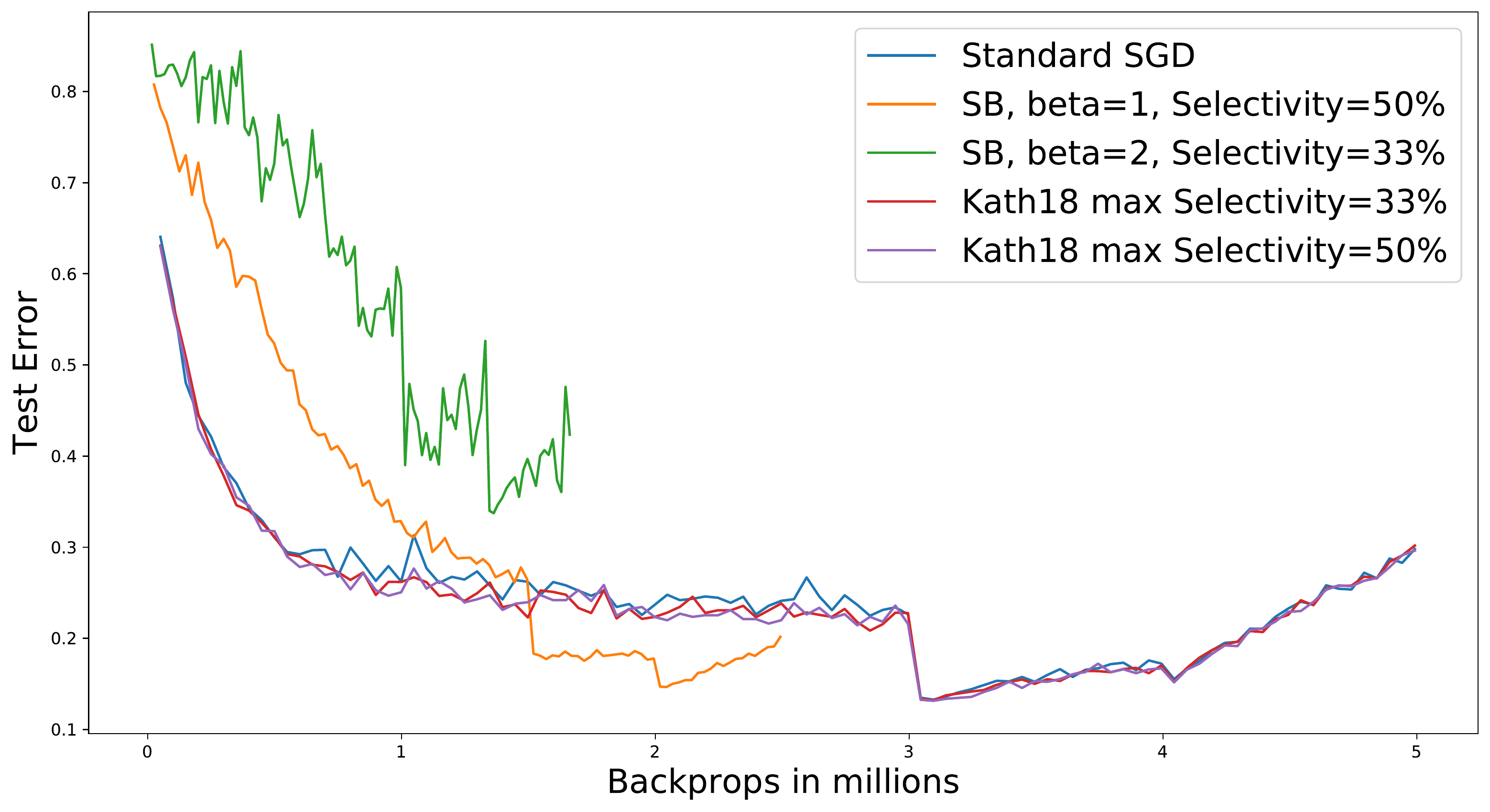}
    	\caption{Corrupted Labels (50 $\%$) \label{fig:50noise}}
	\end{subfigure}
	\begin{subfigure}
		\centering
    	\includegraphics[scale=0.35]{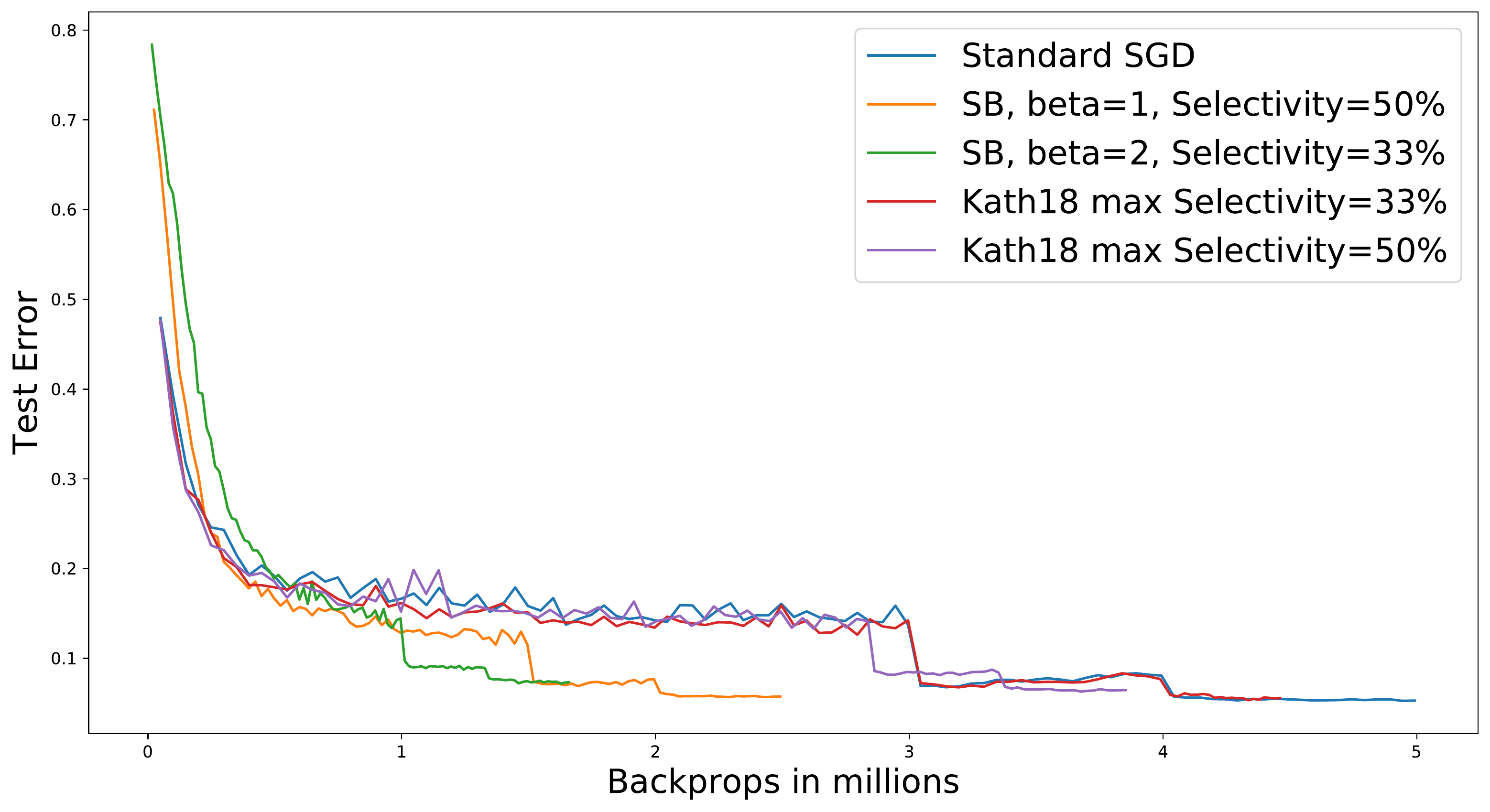}
    		\caption{Gaussian (25 $\%$) \label{fig:25gaussian}}
	\end{subfigure} 
	\begin{subfigure}
		\centering
    	\includegraphics[scale=0.35]{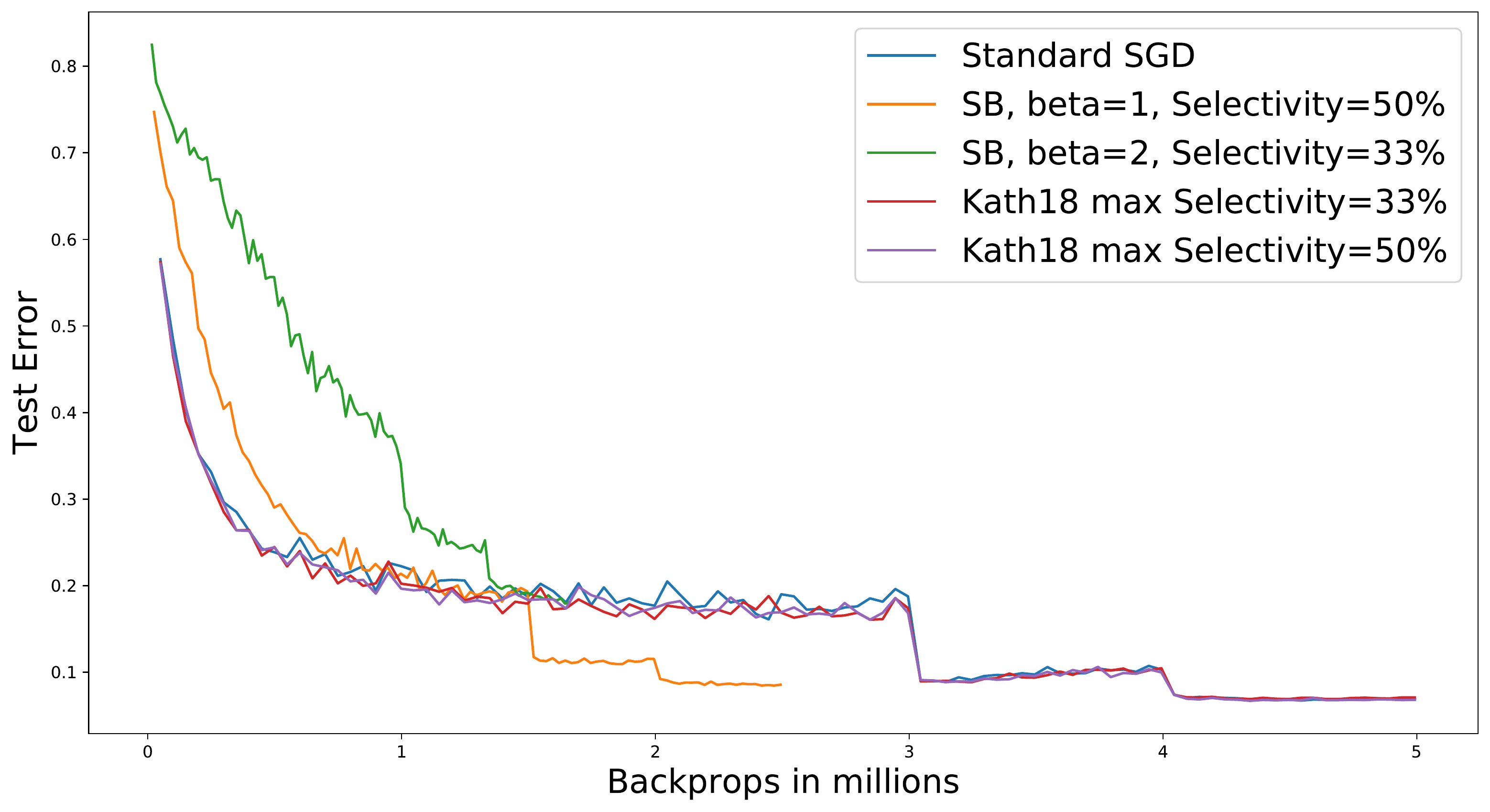}
    	\caption{Gaussian(50 $\%$) \label{fig:50gaussian}}
	\end{subfigure}
\end{figure*}

\begin{figure*}[ht!]
\centering
	\begin{subfigure}
		\centering
    	\includegraphics[scale=0.35]{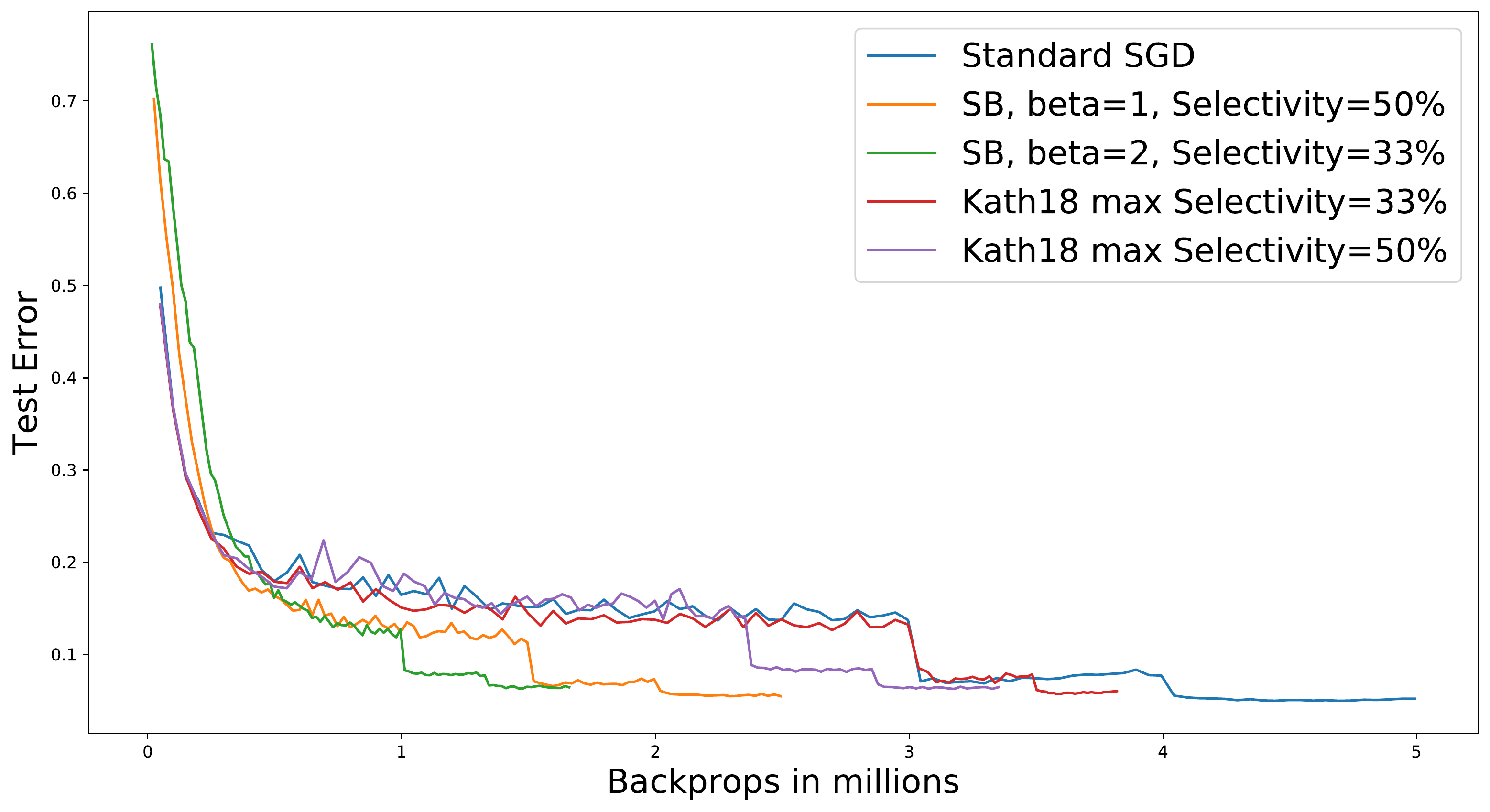}
    	\caption{Shuffle (25 $\%$) \label{fig:25shuffle}}
	\end{subfigure} 
	\begin{subfigure}
		\centering
    	\includegraphics[scale=0.35]{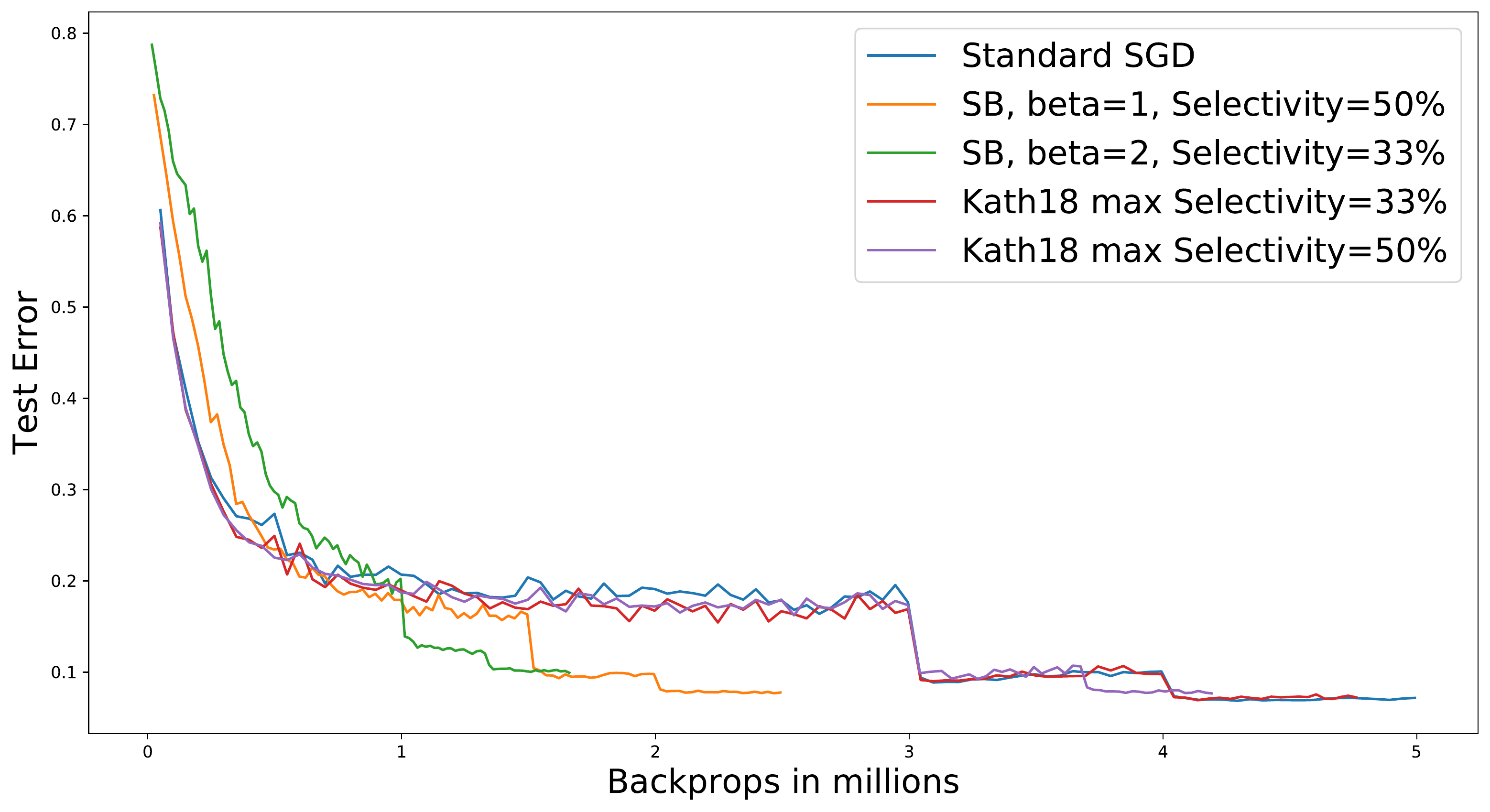}
    	\caption{Shuffle (50 $\%$)  \label{fig:50shuffle}}
	\end{subfigure}
\end{figure*}

In \ref{fig:no corruption} we show the performance of the methods under consideration on a dataset without any corruption.
In \ref{fig:25noise} to \ref{fig:50shuffle}, we depict the performance in the face of various types and amounts of corruption.

\vskip -1in

\end{document}